\documentclass[
]{ceurart}

\usepackage{listings}
\usepackage{subcaption}
\usepackage{overpic}
\usepackage{xcolor}
\usepackage{todonotes}

\begin{document}

\copyrightyear{2026}
\copyrightclause{Copyright for this paper by its authors.
  Use permitted under Creative Commons License Attribution 4.0
  International (CC BY 4.0).}

\conference{Ital-IA 2026: 6th National Conference on Artificial Intelligence, organized by CINI, June 18-19, 2026, Rome, Italy}

\title{xperception - Making Robotic Grasping Easier}

\author[]{Matteo Bortolon}[%
% orcid=0009-0006-2227-6931,
email=mbortolon@fbk.eu,
%url=,
]

\author[]{Andrea Caraffa}[%
% orcid=0009-0000-4687-4297,
email=acaraffa@fbk.eu,
%url=,
]

\author[]{Alice Fasoli}[%
% orcid=0009-0001-5919-9515,
email=alfasoli@fbk.eu,
%url=,
]

\author[]{Fabio Poiesi}[%
% orcid=0009-0000-1840-436X,
email=poiesi@fbk.eu,
%url=,
]
\cormark[1]

\address[]{Fondazione Bruno Kessler, Trento, Italy}

\cortext[1]{Corresponding author.}

\begin{abstract}
The transition toward high-mix low-volume manufacturing demands flexibility in robotic manipulation. 
However, conventional vision systems remain a bottleneck, requiring extensive data collection and model retraining whenever a new object is introduced to the production line. 
To overcome this rigidity, we present xperception\footnote{\url{https://xperception.ai}, accessed: May 2026.}, a zero-shot 6D pose estimation technology that eliminates the need for object-specific fine-tuning and laborious data annotation. 
By directly utilizing typical CAD models and integrating the rich semantic features of foundation models (e.g.~DINOv2, GeDi), xperception achieves millimeter-accurate 6D pose estimation.
xperception showed robustness against severe occlusions in industrial tasks like bin picking and is engineered for deployment on industrial edge hardware, such as NVIDIA Jetson Thor.
Validated at a TRL of 6, the core methodology behind xperception is based on the FreeZe algorithm, which won the international BOP Challenge 2024, paving the way for scalable, plug-and-play robotic automation in unstructured high-mix low-volume manufacturing industries.
\end{abstract}

\begin{keywords}
High-mix low-volume industry \sep
zero-shot learning \sep
vision-based object 6D pose estimation
\end{keywords}

\maketitle

%%%%%%%%%%%%%%%%%%%%%%%%%%%%%%%%%%%%%%%%%%%%%%%%%%%%%%%%%%%%%%
%%%%%%%%%%%%%%%%%%%%%%%%%%%%%%%%%%%%%%%%%%%%%%%%%%%%%%%%%%%%%%
\section{Introduction}\label{sec:intro}

Industrial automation is rapidly transitioning from a paradigm of mass production, characterized by high volumes of a few products to mass customization, aka \textit{high-mix low-volume} \cite{robotics_ifr_2025}. 
Annual robot installations reached 542,076 units globally, pushing the worldwide operational stock to over 4.6 million units \cite{robotics_ifr_2025}. 
Global robot density has risen to 177 robots per 10,000 manufacturing employees. 
While this growing footprint of automation is key to counteract widespread labor shortages, it exposes the inherent rigidity of current robotic manipulation systems in highly variable real-world environments.
The primary bottleneck lies in the vision systems that guide these robots. 
When a new object is introduced to the production line, conventional vision systems fail to manage the variability; they require long reconfiguration time, data collection, and retraining before the robot can successfully recognize and grasp it. 
This results in high setup times that hinder scalability.

The demand for flexible automation is increasing as robotics expands beyond traditional strongholds into high-variability domains like retail, pharmaceuticals, and food industry. 
As the IFR World Robotics report highlights \cite{robotics_ifr_2025}, these sectors are characterized by product variability where rigid automation frameworks struggle.
As a result, core tasks like bin picking become complex. 
Rather than handling continuous runs of identical components, robots must dynamically identify novel parts, determine their exact 6D orientation, and calculate collision-free grasping trajectories on the fly. 
This high-mix reality demands a level of perceptual adaptability that traditional vision systems cannot provide.

State-of-the-art pose estimation methods for novel objects still require dedicated setups \cite{pose_survey_2026}. 
To overcome these limitations, research has shifted toward leveraging foundation models for unseen object localization \cite{freeze_eccv_2024}. 
By exploiting large-scale pre-training across vast datasets, these models extract visual features that are designed for zero-shot generalization without needing task-specific retraining \cite{dinov2_tmlr_2025}.

In this paper, we present \textbf{xperception}, an innovative zero-shot 6D pose estimation pipeline developed at the Technologies of Vision laboratory within Fondazione Bruno Kessler (FBK)\footnote{\url{http://tev.fbk.eu}, accessed: May 2026.}.
Rooted in the laboratory's expertise in 3D vision \cite{geoze_cvpr_2024, perla_cvpr_2025} and robot learning \cite{freegrasp_iros_2025, grasplat_iros_2025, unograsp_cvpr_2026}, xperception is specifically designed to make robotic grasping adaptable. 
Rather than relying on model fine-tuning or laborious data annotation, our system directly utilizes CAD models to estimate the position and orientation of objects. 
xperception is based on the award-winning FreeZe algorithm \cite{freeze_eccv_2024, freezev2_arxiv_2025} that integrates foundation models like DINOv2 \cite{dinov2_tmlr_2025} and GeDi \cite{gedi_tpami_2023} to extract point-level features for point cloud registration.
In particular, FreeZe outperformed a number of leading international competitors winning the BOP Challenge 2024 \cite{bop2025}.
xperception has also been extensively validated in industrial  scenarios, reaching a TRL close to 6. 
Lastly, we tested it on industrial hardware, including NVIDIA Jetson Thor edge computing modules, showing high uptime in production environments.

%%%%%%%%%%%%%%%%%%%%%%%%%%%%%%%%%%%%%%%%%%%%%%%%%%%%%%%%%%%%%%
%%%%%%%%%%%%%%%%%%%%%%%%%%%%%%%%%%%%%%%%%%%%%%%%%%%%%%%%%%%%%%
\section{Our solution}\label{sec:solution}

xperception is designed to bridge the gap between unstructured physical environments and the rigid accuracies and precisions required by industrial robotic manipulators. 
Unlike conventional vision systems that require large data collection, manual annotation, and neural network retraining for novel objects introduced to a production line, xperception operates on a zero-shot paradigm. 
xperception requires zero prior visual training data of the physical object; it only requires the 3D CAD model.

During operation, xperception ingests visual data from industrial 3D sensors or RGB-D cameras.
This data is processed through our proprietary algorithmic architecture, FreeZe \cite{freezev2_arxiv_2025}, which integrates robust semantic understanding of visual and geometric foundation models \cite{dinov2_tmlr_2025, gedi_tpami_2023}. 
Instead of relying on brittle, task-specific features, xperception leverages semantic representations extracted by the foundation model that are robust to unseen data. 
By matching the features of the scene against the features of the provided CAD model, the system calculates the object's exact 6D pose, i.e.~its X, Y, and Z spatial coordinates (3D translation) alongside its roll, pitch, and yaw orientations (3D rotation).

This methodology yields several important industrial advantages. 
First, it achieves millimeter-level accuracy while executing on conventional edge hardware (such as NVIDIA Jetson Thor modules). 
Second, point-level feature matching enables robustness against occlusions, varying lighting conditions, and sensor noise, which are typical in traditional bin-picking scenarios. 
Lastly, the computed 6D pose is mapped to pre-defined grasping coordinates on the CAD model, allowing the system to generate deterministic, collision-free picking trajectories that are fed directly to the robot controller.

%%%%%%%%%%%%%%%%%%%%%%%%%%%%%%%%%%%%%%%%%%%%%%%%%%%%%%%%%%%%%%
\subsection{Industrial Validation: Automatica 2025 Demonstrator}

To validate the maturity, accuracy, precision, and plug-and-play nature of this approach, xperception was deployed in a challenging manipulation task at the Automatica 2025 trade fair. 
Developed in collaboration with the system integrator Metaup srl, the technology was showcased at the Peitian Robot booth. 
The demonstrator featured a robotic manipulator tasked with picking unstructured, randomly scattered pens from a planar surface and feeding them into a laser marking printer.

This specific application highlights the necessity of accurate 6D pose estimation in manufacturing. 
During the demonstration, trade fair visitors could request personalized text to be engraved onto a pen. 
For the laser printer to execute this task correctly, the text had to be aligned in one exact location on the pen's curved surface. 
Therefore, the robot could not simply pick up the pen blindly, it had to grasp the object and place it into the marking machine with near-zero translational and rotational error.

% --------------------------
\begin{figure}[t]
    \centering
    \begin{overpic}[width=1\textwidth]{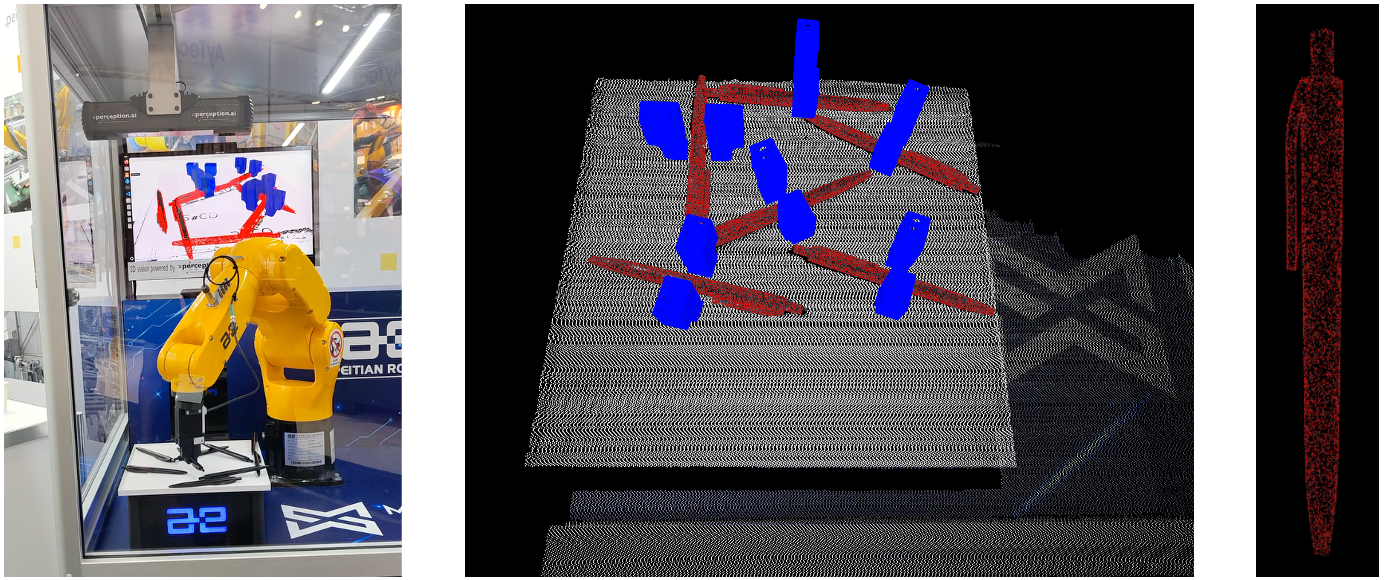}
        \put(9, -2){\footnotesize Robotic system}
        \put(37, -2){\footnotesize 6D pose estimation result + gripper poses on point cloud}
        \put(93, -2){\footnotesize CAD}
    \end{overpic}
    \vspace{.5mm}
    \caption{Visualization of the xperception output during the Automatica 2025 demonstrator. The robotic system leverages the object's CAD model to estimate the 6D pose of randomly scattered pens, identifying optimal, collision-free grasping points (indicated in blue) for the robotic gripper.
    }
    \label{fig:automatica_demo}
\end{figure}
% --------------------------

Our sensory setup utilized a Photoneo MotionCam-3D (Model S)\footnote{\url{www.photoneo.com}, accessed: May 2026.} to capture point clouds of the scene. 
Setup for the task required zero model training; we simply uploaded the CAD model of the pen into the xperception system and pre-defined optimal grasping points directly on the digital model (illustrated as the blue markers in Fig.~\ref{fig:automatica_demo}). 

As the process ran, the vision system repeatedly analyzed the random pile of pens. 
Despite the visual uniformity of the objects and the presence of mutual occlusions, xperception reliably identified individual pens, extracted their 6D pose, and relayed this data to the Peitian manipulator. 
The robot then executed the grasp and aligned the pen within the laser marker socket. 
This demonstrator successfully showed that xperception can eliminate long setup times, transforming a complex, high-precision mechanical task into a robust, easily deployable software solution.

%%%%%%%%%%%%%%%%%%%%%%%%%%%%%%%%%%%%%%%%%%%%%%%%%%%%%%%%%%%%%%
%%%%%%%%%%%%%%%%%%%%%%%%%%%%%%%%%%%%%%%%%%%%%%%%%%%%%%%%%%%%%%
\section{Conclusions}\label{sec:conclusions}

The future of industrial automation demands the high-mix low-volume production paradigm \cite{robotics_ifr_2025}. 
As long as robotic manipulators remain bottlenecked by rigid, data-hungry computer vision systems, the true potential of flexible manufacturing cannot be realized. 
We presented how xperception can effectively shatter this barrier. 
By shifting from object-specific model retraining to a zero-shot, CAD-driven architecture empowered by visual foundation models, xperception delivers millimeter-accurate 6D pose estimation without requiring a single piece of annotated training data. 

Our validations, ranging from setting state-of-the-art benchmarks at the international BOP Challenge 2024 to executing precision tasks on industrial edge hardware, show that xperception is both scientifically advanced and industrially robust.
xperception aims to providing raw visual data into collision-free robotic trajectories, minimizing setup times while ensuring reliability even under severe occlusions.
xperception aspires to become the perceptual engine required to unlock scalable, plug-and-play robotic manipulation for the high-mix low-volume industrial environments of tomorrow.

\section*{Declaration on Generative AI}
 During the preparation of this work, the authors used Gemini 3.1 Pro in order to: Grammar and spelling check; Paraphrase and reword; Improve writing style.  After using these tools/services, the authors reviewed and edited the content as needed and takes full responsibility for the publication’s content.

\bibliography{paper}

\end{document}